\documentclass[twoside,11pt]{article}

\usepackage{blindtext}

\usepackage[preprint]{jmlr2e}

\editor{}
\usepackage{amsmath}
\usepackage{graphicx}
\usepackage{subcaption}
\usepackage{booktabs}

\graphicspath{{media/}}

\usepackage{lastpage}
\jmlrheading{26}{2025}{1-\pageref{LastPage}}{5/25; Revised 5/25}{9/25}{21-0000}{Mannan Bhardwaj}

\ShortHeadings{Interpretable Emergent Language Using Inter-Agent
Transformers}{Bhardwaj}
\firstpageno{1}

\begin{document}

\title{Interpretable Emergent Language Using Inter-Agent
Transformers}

\author{\name Mannan Bhardwaj \email mannanb728@gmail.com \\
       \addr Independant \\
       Hillsborough, NJ 98195-4322, USA}

\maketitle

\begin{abstract}%
This paper explores the emergence of language in multi-agent reinforcement learning (MARL) using transformers. Existing methods such as RIAL, DIAL, and CommNet enable agent communication but lack interpretability. We propose Differentiable Inter-Agent Transformers (DIAT), which leverage self-attention to learn symbolic, human-understandable communication protocols. Through experiments, DIAT demonstrates the ability to encode observations into interpretable vocabularies and meaningful embeddings, effectively solving cooperative tasks. These results highlight the potential of DIAT for interpretable communication in complex multi-agent environments.
\end{abstract}


\section{Introduction}
Human language plays a vital role in human development. It enables the foundation of culture and social interaction: communication, learning, and the exchange of ideas. It is equally important when we study interactions among multiple artificially intelligent agents. More specifically, a compelling area of research is how language emerges between intelligent agents. Rapid progress in machine learning has opened new avenues for exploring such emergent language. For example, how can machine learning models learn to communicate meaningful information to each other? How can this information be used to improve specific tasks?

In order to facilitate language, multi-agent reinforcement learning (MARL) is a promising framework for creating communication-based environments \citep{zhu_survey_2024}. Research in MARL spans from agents working against each other in competitive scenarios to agents working towards one common goal in cooperative environments. Emergent language falls into the cooperation category, where agents try to create a common language autonomously to communicate and collaborate more effectively. In recent years, Transformers \citep{vaswani_attention_nodate} have taken over as the state-of-the-art for language modeling. Self-attention mechanisms allow for more accurate and efficient sequence processing than previous architectures such as CNNs and RNNs \citep{tang_why_2018}. This paper explores the use of transformers in MARL to model emergent language between agents.

Previous works have explored communication via recurrent neural networks. \textit{Reinforced inter-agent learning} (RIAL) \citep{foerster_learning_nodate} uses a recurrent architecture along with independent Q-learning. With RIAL, each agent learns by treating all other agents as a part of the environment. Each agent generates a message as input for other agents in each turn. \textit{Differentiable inter-agent learning} (DIAL) is based on RIAL, but moves away from independent learning. Rather, it takes advantage of the rich gradients that are passed from agent to agent as they communicate. This allows for more stable and efficient learning by essentially treating communication channels as bottleneck layers. This approach uses what is known as \textit{centralized learning}. Centralized learning means that the listening agent passes its gradient back to the speaker agent during training, as opposed to decentralized learning where each agent must learn completely independently of the other agent. This approach also uses \textit{decentralized execution} since each actor can perform actions independently. 

\textit{CommNet} \citep{sukhbaatar2016learningmultiagentcommunicationbackpropagation} aims to solve a similar problem by allowing communication between agents as a method to improve the reward the collective system of agents receives. However, CommNet involves training a single policy for all agents for both actions and communications. It also uses simple multi-layer perceptrons rather than more complex architectures like RNNs. At each turn, it averages all communication states from each agent as inputs for each agent's next turn. 

Unlike DIAL and CommNet \textit{ATOC} \citep{jiang2018learningattentionalcommunicationmultiagent}, it uses a \textit{Deep Deterministic Policy Gradient} (DDPG) to train its parameters. DDPG employs an actor-critic model similar to deep Q networks. CommNet uses an RNN that behaves as an attention unit for deciding what to communicate. Agents in ATOC must choose collaborators based on proximity in the environment. Once collaborators are chosen, a group among those agents is formed. For communication, ATOC uses an LSTM which integrates all states of each agent in a group. ATOC proposes an interesting case where agents are unable to globally communicate and rather can only communicate locally. 

These methods, as well as similar methods \citep{das2020tarmactargetedmultiagentcommunication, kim_communication_2021}, learn communication to solve a MARL task more efficiently. However, their models lack human interpretability since all communication is kept within a high-dimensional, continuous vector space. Harvylov and Titov have explored creating a discrete vocabulary over a wide range of images \citep{havrylov2017emergencelanguagemultiagentgames}. They were the first to show that structured protocols as a string of symbols could be learned via a collaborative task. This structured protocol was learned via LSTMs for sequence-based communication. Mordatch and Abbeel \citep{mordatch2018emergencegroundedcompositionallanguage} show a similar discrete language emergence via Q-learning. Both methods train speaking and listening agents by backpropagating gradients through both agents using Gumbel-softmax \citep{jang2016categorical} to pass gradients through the softmax constriction, thus making both methods centralized learning.

In this paper, we propose a new architecture for creating language: \textit{Differentiable inter-agent transformer} (DIAT). DIAT uses the transformer architecture to model both a speaking and listening agent who interact to create a unique communication scheme to solve particular tasks\footnote{Code available at https://github.com/MannanB/DIAT}. The training method is decentralized where each agent trains and executes independently. Thus, no gradients are passed from the listener to the speaker. Then, we use DIAT to learn a symbolic language in two simple benchmark tasks. Finally, we examine the communication protocols that have been learned. 

\section{Approach}

In this section, the background for our approach is established. Then, the architecture for DIAT is proposed along with its training scheme.

\subsection{Notation and Background}

\textit{Markov Games} are a multi-agent extension of Markov Decision Processes. A markov game is defined as a tuple $ \Gamma = \left(S, \mathcal{N}, \mathcal{A}, r, P, \gamma, \mu\right) $ where there is a finite number of states \textit{{S}}, a finite number of agents $\mathcal{N}$, action sets for each agent $\mathcal{A}_1$, ... ,$\mathcal{A}_\mathcal{N}$, a reward function $r : S \times \mathcal{A}_1 \times \cdots \times \mathcal{A}_n$, a probability transition function $P : S \times \mathcal{A}_1 \times \cdots \times \mathcal{A}_n \to \Delta(S)$, a discount factor $\gamma \in [0, 1)$ and an initial state distribution $\mu \in \Delta(S)$. In our case, each agent receives an observation $o_i$ from the global state ${s \in S}$. Each agent then learns a policy $\pi_i$, which maps each agent's observation to a distribution over its actions. The agent optimizes its policy to maximize their expected discounted returns $R_t=\sum_{t=0}^{n} \gamma^t r_t(s_t, a_{t}^1, \dots, a_{t}^{\mathcal{N}})$

\textit{Policy Gradients}, rather than optimizing based on Q-values, directly adjust the parameters $\theta$ in order to maximize the objective function $L_t(\theta) = \mathbb{E}_{s_t \sim p^\pi, a \sim \pi_\theta} [R_t]$ along the policy gradient $\nabla_\theta L(\theta)$. This uses an actor-critic setup where an actor is trained to interact with the environment while the critic derives the expected returns (the value) from a hidden state of the actor. These expected returns are used to optimize the policy.

\textit{Proximal Policy Optimization} (PPO) \citep{schulman2017proximalpolicyoptimizationalgorithms} is a type of \textit{policy gradient law} that maximizes the function $$L_t^{\text{CLIP} + \text{VF} + \text{S}}(\theta) = \mathbb{E}_t \left[ L_t^{\text{CLIP}}(\theta) - c_1 L_t^{\text{VF}}(\theta) + c_2 S[\pi_\theta](s_t) \right]$$ for each iteration of the agent training loop, where $$L_t^{\text{CLIP}}(\theta) = \mathbb{E}_t \left[ \min \left( r_t(\theta) \hat{A}_t, \text{clip}(r_t(\theta), 1 - \epsilon, 1 + \epsilon) \hat{A}_t \right) \right]$$
$$L_t^{\text{VF}}(\theta) = \left( V_\theta(s_t) - V_t^{\text{target}} \right)^2$$
\textit{S} is the entropy bonus, and $c_1$ and $c_2$ are constants. The advantages $\hat{A}_t$ are given by $$R_t - V_\theta(s_t)$$. The value function $V_\theta(s_t)$ expected returns generated by the critic. Then the entropy bonus encourages exploration within the environment. PPO excels in many fields because of its first-order optimization, simplicity, and stability. Since samples in the experiments are generally small and any kind of long-term memories quickly become useless as the language evolves, PPO's sample efficiency, lack of a replay buffer, and adaptability to MARL make it optimal for this use case.

\textit{Self-attention} splits the input sequence $I$ into \textit{keys}, \textit{queries}, and \textit{values} via parameters $W_k$, $W_q$, and $W_v$ such that $K = W_kI$, $Q = W_qI$, and $V = W_vI$. These are passed through an attention mechanism, such as scaled dot product attention where $Attention(Q, K, V) = softmax\left(\frac{QK^T}{\sqrt{d_k}}\right)V$. To stabilize learning, residual connections are added by $$LayerNorm(Attention(Q, K, V) + I)$$ Self-attention is expanded on via multi-head attention where several different key, value, and query parameters are learned, which are then attended to individually and concatenated to a single matrix in the end.

\subsection{Setup}

There are two agents, a \textit{speaker} and a \textit{listener}, given hyperparameters $(O, V, A)$ where $O$ is the observation sequence size, $V$ is the set of discrete vocabulary, and $A$ is the set of possible actions. The speaker receives a sequence of size $O + C$ where $C$ is the current size of the communication sequence. For each turn, the speaker generates a new communication character $c_i \in V$ which is added to the communication sequence. The augmented communication sequence is passed to the listener which then generates some action $a_i \in A$.

\begin{figure}[ht]
  \centering
  \includegraphics[width=350pt]{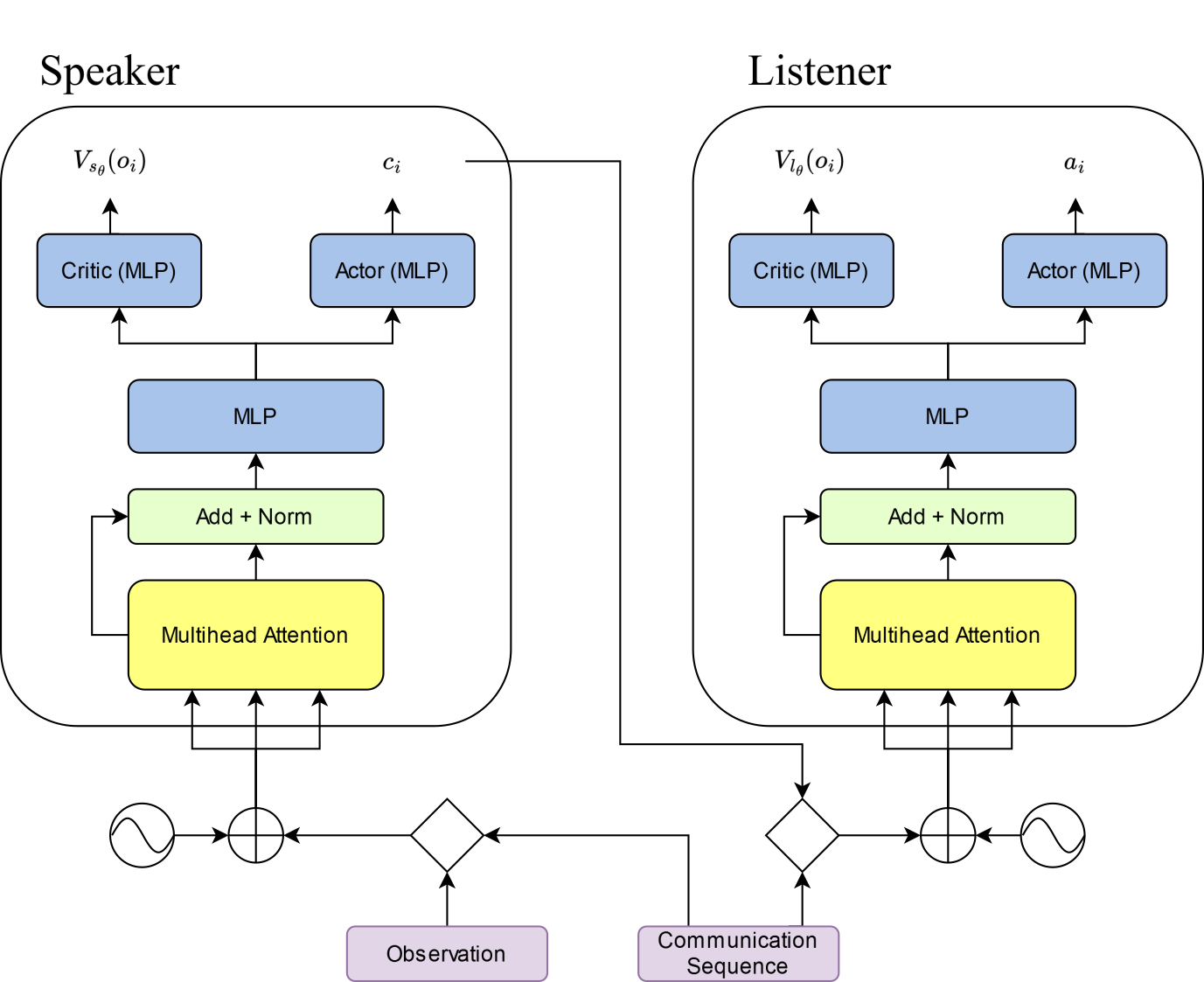}
  \caption{Overall Architecture of Communication Scheme}
  \label{fig:fig1}
\end{figure}

Figure \ref{fig:fig1} outlines the architecture of the proposed communication scheme. The observation $o_i$ and current communication sequence $C_i$ are encoded with both an embedding layer and Rotary Positional Encoding (RoPE) \citep{sukhbaatar2016learningmultiagentcommunicationbackpropagation}. The encoded input is then passed through multi-head self-attention. After a normalization layer, several multi-layer perceptron (MLP) layers output both the speaker's value $V_{s\theta}(o_i)$ for this observation and the logits for the character in the communication sequence where $argmax(softmax(Actor_{mlp}))$ discretizes the logits into a single character $c_i \in V$. $c_i$ is appended to the communication sequence to create $C_{i+1}$. Now, this new communication sequence is passed to the listener model. Note that the listener contains completely unrelated weights, including its embeddings. Thus, during training, it must come up with its own ideas on what various symbols must mean based on the rewards. The listener model has an architecture identical to the speaker except that its only input is $C_{i+1}$. The listener produces the action $a_i$ and the listener's value $V_{l\theta}(o_i)$.

PPO is used to optimize both the speaker and the listener. Each step is recorded in such a way that the observation for the speaker is a combination of the actual environment observation and the current communication sequence, and the observation for the listener is only the current communication sequence. The losses $L_t^{speaker}(\theta)$ and $L_t^{listener}(\theta)$  are all calculated independently for both the speaker and the listener, although the reward $R_t$ is the same for both. Stochastic gradient descent is applied to the total loss $L_t^{total}(\theta) = L_t^{speaker}(\theta) + L_t^{listener}(\theta)$. Each rollout, including the communication sequence, the observation, and reward at each step, is recorded and saved to memory. In every $n$ rollouts, both the speaker and the listener are optimized for multiple SGD steps.  

\section{Experiments}

In this section, DIAT is used to create symbolic language. The embedding layers of both the speaker and listener are isolated for analysis. The accuracy graphs show the accuracy for each 1000 steps during training. In our experiments, the hyperparameters vary for stability. However, in general, the learning rate is confined to $(5 * 10^{-5}, 1 * 10^{-4})$, and the discount factor is $\lambda \in (.95, .99)$. For PPO, the clipping value is $\epsilon = 0.15$. The Adam optimizer \citep{kingma2017adammethodstochasticoptimization} was used to optimize the PPO policy. 

\subsection{Simple Symbolic Language}

In this experiment, the speaker is given a single observation $o_i \in {A, B, \dots}$ and is allowed to communicate via characters $c_i \in {a, b, \dots}$ for a maximum of $t_s$ timesteps. Note that $o_i$ is the same for every $i \in t_s$. This is also known as a \textit{referential game} \citep{lewis2008convention} where the speaker attempts to communicate its observation to the listener, which has to predict what the speaker's observation was based on a low-bandwidth communication channel. The total possible observations $n_o$ and the communication characters $n_c$ vary. The listener has an action space equal to that of the total possible observations. At each time step, given $o_i$ and $a_i$,
\begin{equation}
    r_i = \begin{cases} 
\frac{(t_s + 1)}{2} & \text{if } o_i = a_i, \\
-(i / t_s) & \text{if } o_i \neq a_i, \\
-2 & \text{if } i > t_s
\end{cases}
\end{equation}
The reward function discourages the model from taking too long to get to an answer. The rollout ends when $i > t_s$ or $o_i = a_i$. Thus, the model receives the most reward from guessing $o_i$ correctly when $i = 1$, but the least reward if the model never guesses correctly when $i > t_s$.

\begin{figure}[ht]
    \centering
    \begin{subfigure}[t]{0.20\textwidth}
        \centering
        \caption{Emergent Vocabulary $(n_o = 4, n_c = 2, t_s = 4)$}
        \label{tab:trial1}
        \vspace{0.5em}
        
        \begin{tabular}{l|l|l|l}
            A & b  & C & a  \\
            B & bb & D & ba
        \end{tabular}
        
        \vspace{0.8em} 
        
        \includegraphics[scale=0.3]{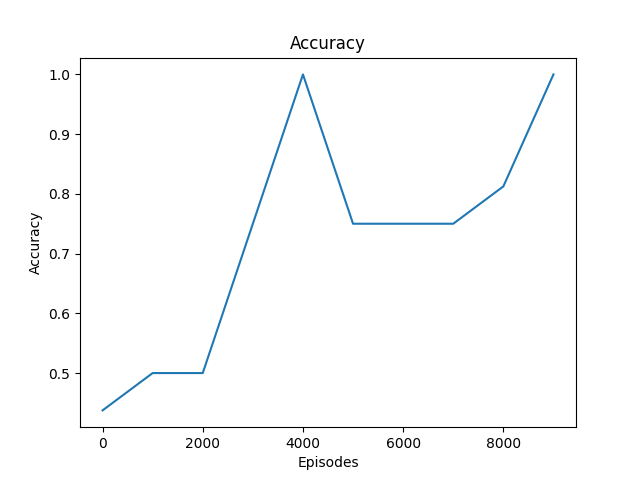}
        \caption*{Accuracy graph of Trial 1}
        \label{fig:fig2}
    \end{subfigure}
    \hfill
    \begin{subfigure}[t]{0.20\textwidth}
        \centering
        \caption{Emergent Vocabulary $(n_o = 10, n_c = 3, t_s = 4)$}
        \label{tab:trial2}
        \vspace{0.5em}
        
        \begin{tabular}{l|l|l|l}
            A & cc  & F & n/a \\
            B & bbb & G & c   \\
            C & b   & H & a   \\
            D & cbb & I & aa  \\
            E & cb  & J & ba
        \end{tabular}
        
        \vspace{0.8em}
        
        \includegraphics[scale=0.3]{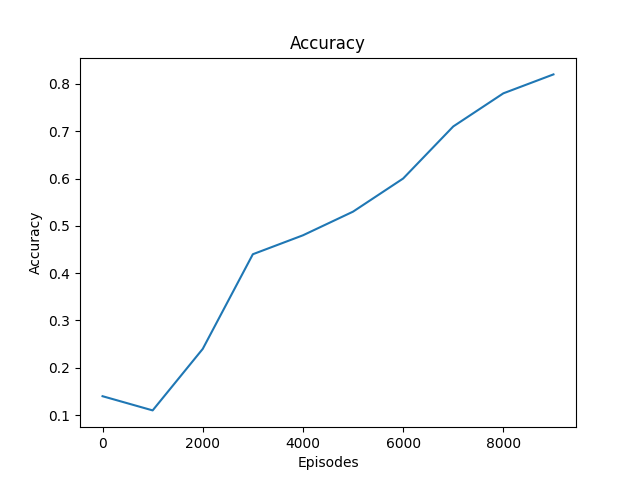}
        \caption*{Accuracy graph of Trial 2}
        \label{fig:fig3}
    \end{subfigure}
    \hfill
    \begin{subfigure}[t]{0.20\textwidth}
        \centering
        \caption{Emergent Vocabulary $(n_o = 26, n_c = 3, t_s = 10)$}
        \label{tab:trial3}
        \vspace{0.5em}
        
        \begin{tabular}{l|l|l|l}
            A & bcccc     & O & babccc \\
            D & ccccccccb & P & bcc    \\
            R & bccccccc  & Z & ab    
        \end{tabular}
        
        \vspace{0.8em}
        
        \includegraphics[scale=0.3]{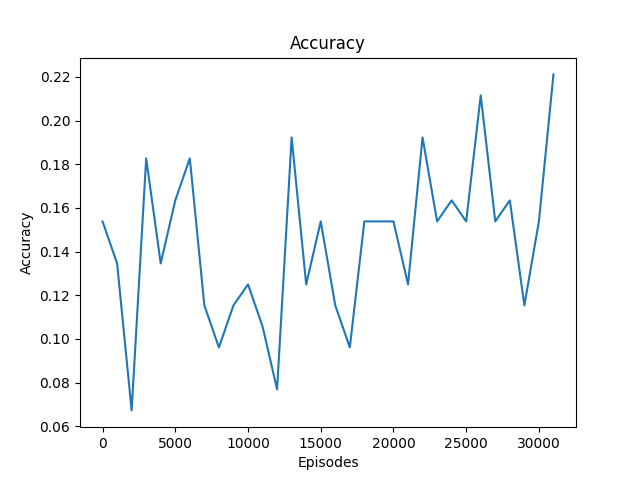}
        \caption*{Accuracy graph of Trial 3}
        \label{fig:fig4}
    \end{subfigure}

    \caption{Tables and accuracy graphs for Trials 1--3 }
    \label{fig:combinedTrials}
\end{figure}

\textbf{Trial 1} The results are shown in Figure \ref{fig:fig2}. Here, a very simple and constrained environment was tested with only 4 possible observations and a two-character vocabulary. $t_s$ is 4. However, none of the vocabulary became that long due to the discouragement of long rollouts. Generally, training was stable and the model easily came to these conclusions. 

\textbf{Trial 2} The results are shown in Figure \ref{fig:fig3}. With $n_o = 10, n_c = 3$, and $t_s = 4$, all except one observation was represented correctly.

\textbf{Trial 3} The results are shown in Figure \ref{fig:fig4}. The vocabulary space and $t_s$ were expanded dramatically. The model only managed to create consistent representations for a small portion of the observation space. Nonetheless, with the trajectory of the accuracy graph, more training might have allowed the model to converge eventually. 

The model was able to come up with unique sequences of vocabulary to represent each unique observation. Yet, in certain observations, no suitable sequence converged. This was especially true as $n_c$ and $n_o$ were increased, as can be seen in Trial 3. Trials 1 and 2, however, achieved satisfactory results. A useful, human-understandable, communication language was established.

\subsection{Encoding Meaning Using Shapes and Colors}

\begin{figure}[ht]
  \centering
  \includegraphics[width=0.75\textwidth]{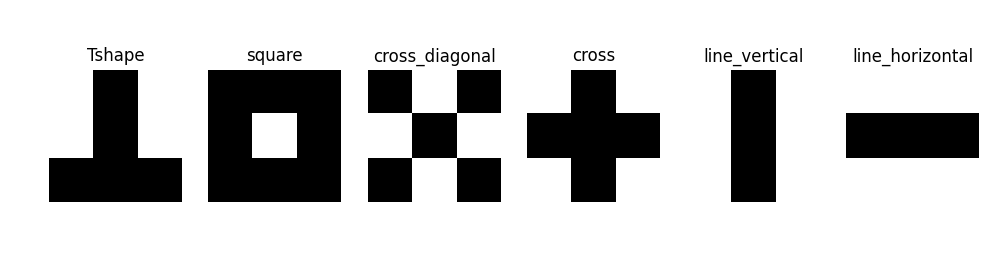}
  \caption{Shapes in Observation Space.}
  \label{fig:shapes}
\end{figure}

In this experiment, the speaker is given a $3\times3$ shape (see Figure \ref{fig:shapes}) and a color (red or blue). The listener needs to predict the correct color and shape. The shape observation is $o_i^s$ and the color observation is $o_i^c$. The listener outputs a prediction of the shape $a_i^s$ and a prediction of the color $a_i^c$. The reward model is as follows:
\begin{equation}
    r_i = \begin{cases} 
\frac{(t_s + 1)}{2} & \text{if } o_i^s = a_i^s \text{ and } o_i^c = a_i^c, \\
1 & \text{if } o_i^s = a_i^s \text{ and } o_i^c \neq a_i^c, \\
0 & \text{if } o_i^s \neq a_i^s \text{ and } o_i^c = a_i^c, \\
-(i / t_s) & \text{if }  o_i^s \neq a_i^s \text{ and } o_i^c \neq a_i^c, \\
-2 & \text{if } i > t_s
\end{cases}
\end{equation}

This is a similar setup to the first experiment. There is an increased reward if the listener gets either $a_i^c$ or $a_i^s$ correct, with $a_i^s$ being the larger reward since there are more shapes to choose from. The shapes are flattened to a sequence of length 9 to be passed through the attention mechanism. The empty space is 0, red is 1, and blue is 2. Then, the entire observation is shifted by $n_c$. This experiment only included one trial where $n_o = 6 \times 2 = 12$, $n_v = 6$, and $t_s = 4$. 

\begin{table}[ht]
\centering
\begin{tabular}{ll}
\toprule
\textbf{Color} & \textbf{Learned Vocabulary Sequences} \\
\midrule
Red  & \texttt{ed}, \texttt{ee}, \texttt{b}, \texttt{be}, \texttt{eee}, \texttt{bbb}, \texttt{d}, \texttt{e} \\
Blue & \texttt{f}, \texttt{ca}, \texttt{aa}, \texttt{cac}, \texttt{cabc}, \texttt{c}, \texttt{ff}, \texttt{a}, \texttt{ccc} \\
\bottomrule
\end{tabular}
\caption{Vocabulary corresponding to color only.}
\label{tab:colorvocab}
\end{table}

\begin{table}[ht]
\centering
\begin{tabular}{ll}
\toprule
\textbf{Shape} & \textbf{Learned Vocabulary Sequences} \\
\midrule
T Shape         & \texttt{eee}, \texttt{ee}, \texttt{aa} \\
Cross           & \texttt{be}, \texttt{caa}, \texttt{ca} \\
Square          & \texttt{f}, \texttt{ffd}, \texttt{ff}, \texttt{d}, \texttt{fff} \\
Line Vertical   & \texttt{c}, \texttt{b} \\
Cross Diagonal  & \texttt{e}, \texttt{a} \\
Line Horizontal & \texttt{ccc}, \texttt{bbb}, \texttt{cc}, \texttt{bb} \\
\bottomrule
\end{tabular}
\caption{Vocabulary corresponding to shape only.}
\label{tab:shapevocab}
\end{table}

\begin{table}[ht]
\centering
\begin{tabular}{lllllll}
\toprule
       & \textbf{T} & \textbf{Square} & \textbf{Cross Diag} & \textbf{Cross} & \textbf{Line V} & \textbf{Line H} \\
\midrule
\textbf{Red}  & \texttt{eee}, \texttt{ee} 
     & \texttt{d} 
     & \texttt{e} 
     & \texttt{be} 
     & \texttt{b} 
     & \texttt{bbb}, \texttt{bb} \\
\textbf{Blue} & \texttt{aaa}, \texttt{aa} 
     & \texttt{f} 
     & \texttt{a} 
     & \texttt{ca} 
     & \texttt{c} 
     & \texttt{ccc}, \texttt{cc} \\
\bottomrule
\end{tabular}
\caption{Vocabulary corresponding to both shape and color.}
\label{tab:shapecolorvocab}
\end{table}

Tables \ref{tab:colorvocab}, \ref{tab:shapevocab}, and \ref{tab:shapecolorvocab} show the learned vocabulary organized by either color, shape, or both. The figures show that the model seems to have learned to encode the colors into the types of characters it uses, and the shape with the actual sequence. In fact, Table \ref{tab:shapecolorvocab} shows how the model has learned to make one character the counterpart of another: $e \leftrightarrow a, d \leftrightarrow f, b \leftrightarrow c$. For example, the red T-shape can either by represented by $eee$ or $ee$. Since the blue counterpart of $e$ is $a$, the blue T-shape is $aaa$ or $aa$. Another interesting emergent behavior was the inclusion of synonyms, where multiple sequences corresponded to one. However, this could also have been a byproduct of shorter-sequence bias. The model might have been in the middle of learning a shorter representation ($eee \rightarrow ee$) as that would achieve a greater reward. If more time had been spent training, perhaps the longer vocabulary would have disappeared. 

Finally, some loose semantic meanings of each shape have been learned. For example, the vertical and horizontal lines are very similar shapes, only differing by rotation. This fact seems to have been encoded within the vocabulary since both the vertical line and horizontal line actually use the same characters and only differ by the number of them. Similarly, the word for Cross is simply the combination of a Vertical Line and a Diagonal Cross, which makes sense since the normal cross is a "straighter" diagonal cross.

\begin{figure}[ht]
  \centering
  \includegraphics[scale=0.6]{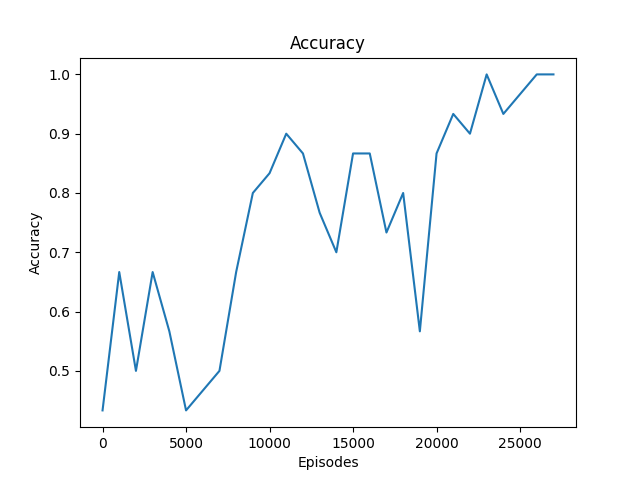}
  \caption{Accuracy curve during training for the shapes-and-colors experiment.}
  \label{fig:accuracy}
\end{figure}

\begin{figure}[ht]
  \centering
  \includegraphics[scale=0.6]{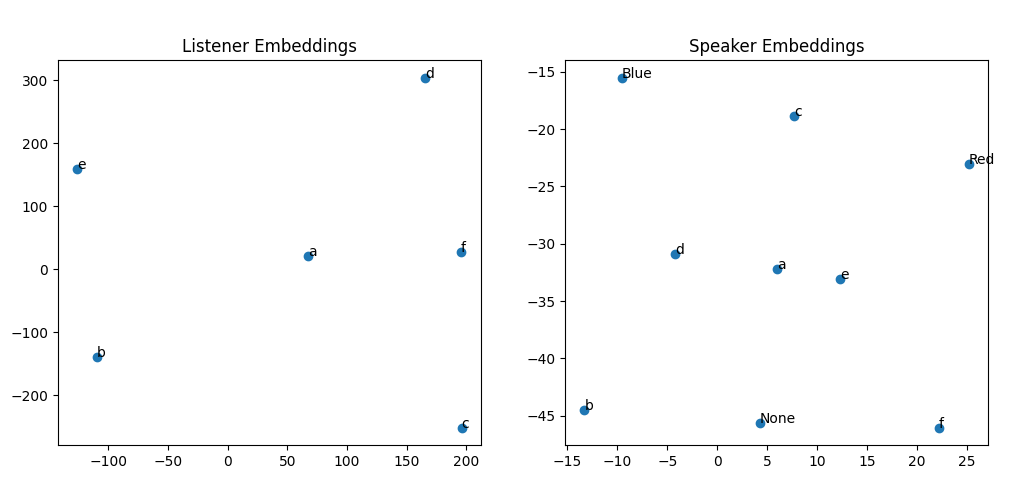}
  \caption{t-SNE of the learned embeddings for the speaker and listener.}
  \label{fig:embeddings}
\end{figure}

Analyzing the learned embeddings shows that some meaning has actually been conveyed. The embedding layer encodes each character to some learned high-dimension vector representation. t-NSE \citep{van2008visualizing} was used to map this higher-dimension vector to a 2D space for analysis. Figure \ref{fig:embeddings} shows this embedding map for both the speaker and the listener. Recall that the embedding layers were trained separately. The listener embeddings show that the "red" characters ($b$, $e$, and $d$) are closer together and are on their side compared to that of the "blue" ($a$, $c$, and $f$) characters. However, this isn't reflected in the speaker embeddings. In fact, the color relations are not represented at all. This could be due to the fact that the speaker embeddings do not need to encode specific color information since it is directly given to it in its observation. In other words, since it knows that it needs to communicate "red," it does not care about the relationship between "red" and $a$ within its embeddings. 

\subsection{Spatial language and understanding}

\begin{figure}[h]
  \centering
  \includegraphics[scale=0.6]{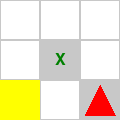}
  \caption{Sample of a state of the environment. The yellow square is the rescuer, the x is the survivor, and the triangle is the obstacle.}
  \label{fig:grid}
\end{figure}

In this section, DIAT is used to come up with a communication scheme to communicate the location of an object in a grid space. The environment is loosely based on a scenario where there has been some kind of disaster, and one or more survivors are scattered around a grid. A rescuer is placed on the grid with no visibility of its surroundings while an outside party is communicating where each of the survivors are. The grid (Figure \ref{fig:grid}) is a square grid with side length $l_g$ where there is one survivor as well as $n_o$ obstacles. The observer can see where all the survivors and obstacles are, but not where the rescuer is. The rescuer can see the observer's communication sequence as well as its location. For each rollout, the observer generates a single communication string of length $l_c$ on turn 1, which is passed to the rescuer for the rest of the turns. The rescuer can then move through the grid to find the survivors. The game ends once the rescuer is on the same tile as a survivor or obstacle or when the agent exceeds the maximum number of time steps $t_s$.

For all experiments, $l_c=3$, $n_c=3$, $t_s=4$, and $l_g=3$ (the length of the communication string is 3, the size of the vocabulary is 3, each rollout is a maximum of 4 turns, and the side length of the square grid is 3). This means that there are $3^3=27$ unique communication sequences possible. At each time step: $r_i = 2$ when moving towards a survivor, $r_i = -4$ when moving away from a survivor, $r_i = 15$ when the rescuer finds a survivor, and $r_i=-20$ either when the rescuer runs out of time (timesteps exceed 4) or when the rescuer hits an obstacle. 

\begin{figure}[h]
  \centering
  \includegraphics[scale=0.6]{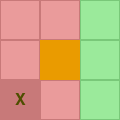}
  \caption{Learned spatial representation without obstacles. CCC $\rightarrow$ red and AAA $\rightarrow$ green}
  \label{fig:grid_colors}
\end{figure}

\textbf{Trial 1} In trial one, $n_o=0$. Therefore, the observer only needs to encode the position of the survivor in its communication string. Theoretically, a perfect solution would map each of the 27 possible positions of the survivor to a unique communication string. However, this environment was found to be much more complex than the previous two. The only way for unique communication strings to occur is if a particular unique sequence randomly happened to correlate to the listening agent finding the survivor in a particular spot slightly more than average. This would cause a feedback effect where the rescuer would learn to match this communication string with a certain position of the survivor more often. This would cause the observer to produce the same string when prompted with a certain position more often. Due to the difficulty and randomness of this task, in Figure \ref{fig:grid_colors}, the observer was only able to communicate two parts of the grid with its communication sequence. The listener is then able to search the area communicated.

\textbf{Trial 2} In trial two, $n_o=1$. Therefore, the observer needed to encode both the position of the obstacle and the survivor in its communication string. A perfect one-to-one map is impossible since there are  $9 \times8=72$ permutations of grid states with only 27 unique communication strings. Thus, some kind of lossy compression is needed to be learned. However, the agent was not able to converge to a stable communication sequence. This is likely caused by how much random chance contributes to both the environment and learning. Since the obstacle is in a random spot every time, the rescuer has an even harder time trying to correlate communication sequences to what is actually on the grid. For example, if the rescuer learned that the communication sequence $AAA$ meant the survivor was in the top-right corner but was unable to learn information about the obstacle, a couple of unlucky run-ins to an obstacle might make the correlation fall apart. 

Table \ref{tab:baseline-diat} shows the baseline and DIAT values for an average of 500 rollouts with one obstacle versus without an obstacle.
\begin{table}[ht]
\centering
\begin{tabular}{lllll}
\toprule

                                                & \multicolumn{2}{c}{\textbf{No Obstacle}}     & \multicolumn{2}{c}{\textbf{One Obstacle}}     \\ \midrule
\multicolumn{1}{l}{\textbf{Average over 500 rollouts}} & \multicolumn{1}{l}{Baseline} & DIAT & \multicolumn{1}{l}{Baseline} & DIAT  \\ \midrule
\multicolumn{1}{l}{Survivors Saved}           & \multicolumn{1}{l}{.34}      & .65  & \multicolumn{1}{l}{.32}      & .39   \\ 
\multicolumn{1}{l}{Reward}                    & \multicolumn{1}{l}{-10.69}   & 2.08 & \multicolumn{1}{l}{-10.26}   & -8.58 \\ 
\multicolumn{1}{l}{Steps Taken}               & \multicolumn{1}{l}{3.38}     & 2.51 & \multicolumn{1}{l}{3.36}     & 3.10  \\ 
\bottomrule
\end{tabular}
\caption{Baseline and DIAT values with and without obstacles}
\label{tab:baseline-diat}
\end{table}

\section{Conclusion}
This paper explored and analyzed emergent language via a novel architecture and MARL. A speaker and listener, each using a transformer architecture, converse with each other to learn a unique language based on the task. The results indicate that an effective and human-understandable language has been produced along with meaningful embeddings which stipulate a deeper understanding of its communication. Our model allows for transparency within complex artificial intelligence environments which builds trust in unsupervised systems. In the future, we intend to extend this model to more complex environments to encode more complex relational data at a larger scale.






\vskip 0.2in
\bibliography{paper}

\begin{thebibliography}{15}
\providecommand{\natexlab}[1]{#1}
\providecommand{\url}[1]{\texttt{#1}}
\expandafter\ifx\csname urlstyle\endcsname\relax
  \providecommand{\doi}[1]{doi: #1}\else
  \providecommand{\doi}{doi: \begingroup \urlstyle{rm}\Url}\fi

\bibitem[Das et~al.(2020)Das, Gervet, Romoff, Batra, Parikh, Rabbat, and Pineau]{das2020tarmactargetedmultiagentcommunication}
Abhishek Das, Théophile Gervet, Joshua Romoff, Dhruv Batra, Devi Parikh, Michael Rabbat, and Joelle Pineau.
\newblock Tarmac: Targeted multi-agent communication, 2020.
\newblock URL \url{https://arxiv.org/abs/1810.11187}.

\bibitem[Foerster et~al.(2016)Foerster, Assael, de~Freitas, and Whiteson]{foerster_learning_nodate}
Jakob~N. Foerster, Yannis~M. Assael, Nando de~Freitas, and Shimon Whiteson.
\newblock Learning to communicate with deep multi-agent reinforcement learning, 2016.
\newblock URL \url{https://arxiv.org/abs/1605.06676}.

\bibitem[Havrylov and Titov(2017)]{havrylov2017emergencelanguagemultiagentgames}
Serhii Havrylov and Ivan Titov.
\newblock Emergence of language with multi-agent games: Learning to communicate with sequences of symbols, 2017.
\newblock URL \url{https://arxiv.org/abs/1705.11192}.

\bibitem[Jang et~al.(2017)Jang, Gu, and Poole]{jang2016categorical}
Eric Jang, Shixiang Gu, and Ben Poole.
\newblock Categorical reparameterization with gumbel-softmax, 2017.
\newblock URL \url{https://arxiv.org/abs/1611.01144}.

\bibitem[Jiang and Lu(2018)]{jiang2018learningattentionalcommunicationmultiagent}
Jiechuan Jiang and Zongqing Lu.
\newblock Learning attentional communication for multi-agent cooperation, 2018.
\newblock URL \url{https://arxiv.org/abs/1805.07733}.

\bibitem[Kim et~al.(2021)Kim, Park, and Sung]{kim_communication_2021}
Woojun Kim, Jongeui Park, and Youngchul Sung.
\newblock Communication in multi-agent reinforcement learning: Intention sharing.
\newblock In \emph{International Conference on Learning Representations}, 2021.
\newblock URL \url{https://openreview.net/forum?id=qpsl2dR9twy}.

\bibitem[Kingma and Ba(2017)]{kingma2017adammethodstochasticoptimization}
Diederik~P. Kingma and Jimmy Ba.
\newblock Adam: A method for stochastic optimization, 2017.
\newblock URL \url{https://arxiv.org/abs/1412.6980}.

\bibitem[Lewis(20082)]{lewis2008convention}
David Lewis.
\newblock \emph{Convention: A philosophical study}.
\newblock John Wiley \& Sons, Ltd, 20082.
\newblock \doi{https://doi.org/10.1002/9780470693711}.

\bibitem[Mordatch and Abbeel(2018)]{mordatch2018emergencegroundedcompositionallanguage}
Igor Mordatch and Pieter Abbeel.
\newblock Emergence of grounded compositional language in multi-agent populations, 2018.
\newblock URL \url{https://arxiv.org/abs/1703.04908}.

\bibitem[Schulman et~al.(2017)Schulman, Wolski, Dhariwal, Radford, and Klimov]{schulman2017proximalpolicyoptimizationalgorithms}
John Schulman, Filip Wolski, Prafulla Dhariwal, Alec Radford, and Oleg Klimov.
\newblock Proximal policy optimization algorithms, 2017.
\newblock URL \url{https://arxiv.org/abs/1707.06347}.

\bibitem[Sukhbaatar et~al.(2016)Sukhbaatar, Szlam, and Fergus]{sukhbaatar2016learningmultiagentcommunicationbackpropagation}
Sainbayar Sukhbaatar, Arthur Szlam, and Rob Fergus.
\newblock Learning multiagent communication with backpropagation, 2016.
\newblock URL \url{https://arxiv.org/abs/1605.07736}.

\bibitem[Tang et~al.(2018)Tang, Müller, Rios, and Sennrich]{tang_why_2018}
Gongbo Tang, Mathias Müller, Annette Rios, and Rico Sennrich.
\newblock Why self-attention? a targeted evaluation of neural machine translation architectures, 2018.
\newblock URL \url{https://arxiv.org/abs/1808.08946}.

\bibitem[van~der Maaten and Hinton(2008)]{van2008visualizing}
Laurens van~der Maaten and Geoffrey Hinton.
\newblock Visualizing data using t-sne.
\newblock \emph{Journal of Machine Learning Research}, 9\penalty0 (86):\penalty0 2579--2605, 2008.
\newblock URL \url{http://jmlr.org/papers/v9/vandermaaten08a.html}.

\bibitem[Vaswani et~al.(2017)Vaswani, Shazeer, Parmar, Uszkoreit, Jones, Gomez, Kaiser, and Polosukhin]{vaswani_attention_nodate}
Ashish Vaswani, Noam Shazeer, Niki Parmar, Jakob Uszkoreit, Llion Jones, Aidan~N. Gomez, \L{}ukasz Kaiser, and Illia Polosukhin.
\newblock Attention is all you need.
\newblock In \emph{Proceedings of the 31st International Conference on Neural Information Processing Systems}, NIPS'17, page 6000–6010, Red Hook, NY, USA, 2017. Curran Associates Inc.
\newblock ISBN 9781510860964.

\bibitem[Zhu et~al.(2024)Zhu, Dastani, and Wang]{zhu_survey_2024}
Changxi Zhu, Mehdi Dastani, and Shihan Wang.
\newblock A survey of multi-agent deep reinforcement learning with communication.
\newblock \emph{Autonomous Agents and Multi-Agent Systems}, 38\penalty0 (1):\penalty0 4, 2024.

\end{thebibliography}

\end{document}